\documentclass[9pt]{article}
\usepackage{spconf,amsmath,graphicx}

\title{ATTENTION-AWARE GENERALIZED MEAN POOLING FOR IMAGE RETRIEVAL}
\name{Yinzheng Gu \qquad Chuanpeng Li \qquad Jinbin Xie}
\address{Video++ AI Lab, Jilian Technology Group, China}
\begin{document}
%
\maketitle
\begin{abstract}
It has been shown that image descriptors extracted by convolutional neural networks (CNNs) achieve remarkable results for retrieval problems. In this paper, we apply attention mechanism to CNN, which aims at enhancing more relevant features that correspond to important keypoints in the input image. The generated attention-aware features are then aggregated by the previous state-of-the-art generalized mean (GeM) pooling followed by normalization to produce a compact global descriptor, which can be efficiently compared to other image descriptors by the dot product. An extensive comparison of our proposed approach with state-of-the-art methods is performed on the new challenging ROxford5k and RParis6k retrieval benchmarks. Results indicate significant improvement over previous work. In particular, our attention-aware GeM (AGeM) descriptor outperforms state-of-the-art method on ROxford5k under the ``Hard'' evaluation protocal.
\end{abstract}

\section{INTRODUCTION}\label{sec:intro}

In computer vision, the task of instance-level image retrieval aims at, given a query image, retrieving all images in a large-scale database that contain the same object as the query. Traditionally, the best performing approaches relied on local invariant features such as SIFT \cite{L2004} and aggregation strategies such as BoW \cite{SZ2003}, VLAD \cite{JDSP2010}, or FV \cite{PLSP2010} built on top of these local features. The resulting representations are usually efficiently indexed and individually matched followed by a re-ranking stage.

Recently, methods based on convolutional neural networks (CNNs) have been advancing rapidly. As a first attempt, several works \cite{BSCL2014, RASC2014, GWGL2014} proposed to use features extracted by fully connected layers and demonstrated satisfactory performance. However, state-of-the-art results were still obtained by non-CNN-based methods and convolutional layers were proposed as an alternative. These feature extractors possess the advantage that a compact representation of fixed length can be produced efficiently from an input image of any size and aspect ratio. Consequently, a series of papers have been written regarding various strategies to construct competitive image representations such as SPoC \cite{BL2015}, CroW \cite{KMO2016}, MAC \cite{TSJ2016}, or R-MAC \cite{TSJ2016} descriptors. These approaches, when combined with proper post-processing techniques, produced for the first time a system that competes or outperforms conventional methods based on local features.

However, a major drawback of the above methods is that image descriptors were directly extracted using off-the-shelf models trained for the classification task. While being efficient, it became evident that the improvement gain was limited due to their lack of learning. The first fine-tuning approach for image retrieval was proposed by Babenko {\em et al.} \cite{BSCL2014} using a classification loss on a new set of landmark images better suited for the retrieval task. Later on, Gordo {\em et al.} \cite{GARL2017} argued that the similarity measure being optimized should coincide with the one to be used during the final task. Building on the R-MAC pipeline, the deep image retrieval (DIR) network was trained end-to-end on a clean version of the Babenko dataset using a ranking loss. Concurrently, the work of Radenovi\'{c} {\em et al.} \cite{RTC2018} dispensed the need of manual effort to collect/clean a large-scale dataset for training by exploiting the structure-from-motion (SfM) pipeline \cite{SRCF2015}. In addition, a novel generalized mean (GeM) pooling operation was introduced offering more performance boost over their previous work \cite{RTC2016}.

Insipired by the recent work of Wang {\em et al.} \cite{WJQYLZWT2017} which inserts attention modules into CNNs to improve performance in the classification task, we incorporate (soft) attention mechanism by considering a two-branch network: The main branch which, same as the above works, consists of the base architecture before the final pooling layer, and an attention branch which consists of additional layers applied to feature maps produced by various prior blocks in the base architecture in a feedforward manner. The outputs of the two branches are then combined together by the attention residual learning mechanism as in \cite{WJQYLZWT2017} to generate attention-aware features which are aggregated by the GeM operation to produce a compact image representation referred to as the attention-aware GeM (AGeM) descriptor. The additional attention branch is easily implementable, trainable via back-propagation, and adds only a small computational overhead.

The rest of this paper is organized as follows. In Section \ref{sec:network}, we describe our proposed network architecture and introduce AGeM descriptor. In Section \ref{sec:experim}, we perform an extensive evaluation comparing various state-of-the-art methods in the literature. The paper concludes with Section \ref{sec:conclu}.

\section{PROPOSED METHOD}\label{sec:network}

\subsection{Network and Pooling}

For our experiments, we choose ResNet-101 \cite{HZRS2016} as our CNN architecture. Given an input image, we take the feature maps produced by the last convolutional layer as output, which is of the form ${\cal X} \in {\bf R}^{W \times H \times K}$, where $K$ denotes the number of channels. We always assume ReLU activation is applied. Denote by ${\cal X}_k \in {\bf R}^{W \times H}$ the $k$-th feature map of ${\cal X}$, we apply a pooling operation to produce a number $F_k$ representing ${\cal X}_k$ so that the input image can be represented by the vector $[F_1, \ldots, F_K]^{\text{T}}$. This vector is then $\ell^2$-normalized so as to have unit norm.

Two of the simplest pooling methods are the average and max pooling operations corresponding to SPoC \cite{BL2015} and MAC \cite{TSJ2016} descriptors, respectively, which already achieve competitively good results on standard benchmarks. In order to further boost the performance, the generalized mean (GeM) pooling was used in \cite{RTC2018} as a replacement, where the corresponding GeM descriptor is given by
\begin{equation}\label{eqn:gem}
[F^{(\text{GeM})}_1, \ldots, F^{(\text{GeM})}_K]^{\text{T}}, \quad F^{(\text{GeM})}_k = \left(\frac{1}{|{\cal X}_k|}\sum_{x \in {\cal X}_k}x^{p_k}\right)^{\frac{1}{p_k}},
\end{equation}
which generalizes SPoC and MAC by taking $p_k = 1$ and $p_k \to \infty$. More importantly, the GeM pooling is a differentiable operation, and hence the whole network can be trained in an end-to-end fashion.

In \eqref{eqn:gem}, there is a different pooling parameter $p_k$ for each feature mep ${\cal X}_k$. However, one can also use a shared parameter $p$ for all feature maps. Following \cite{RTC2018}, we adopt this simpler option as it achieves slightly better results. Finally, we have $K = 2048$ for ResNet-101 so that each of the aforementioned descriptors is a 2048-D compact image representation.

\subsection{Attention-Aware GeM}

We now describe the construction of our attention-aware GeM (AGeM) descriptor. Given an input image, the first block of the ResNet-101 architecture consists of a $7 \times 7$ convolution followed by a $3 \times 3$ max pooling to produce a feature activations output with channel size $64$. Then there are four more residual blocks, denoted $\{B_2, B_3, B_4, B_5\}$, of $1 \times 1$ and $3 \times 3$ convolutional layers producing feature maps of the same size within each block. For $i \in \{2, 3, 4, 5\}$, denote by $B_{i, j}$ the $j$-th residual unit of $B_i$ and $\mathcal{X}_{i, j}$ the feature maps produced by the last layer of $B_{i, j}$. Note that $\mathcal{X}_{i, j}$ has channel size 256 (respectively, 512, 1024, and 2048) for $i = 2$ (respectively, 3, 4, and 5).

Our network architecture is made of two branches. First, there is the main branch which is exactly same as GeM before the final pooling layer that takes an input image and produces feature maps $\mathcal{X}_{5, 3}$ from $B_{5, 3}$ of ResNet-101. For the attention branch, we add three attention units, denoted Att$1$, Att$2\_1$, and Att$2\_2$, which are applied to feature maps $\mathcal{X}_{4, 23}$, $\mathcal{X}_{5, 1}$, and $\mathcal{X}_{5, 2}$ produced by $B_{4, 23}$, $B_{5, 1}$, and $B_{5, 2}$, respectively. The Att$1$ unit consists of four convolutional layers of kernel size $3 \times 3$, $3 \times 3$, $1 \times 1$, and $1 \times 1$, respectively, with stride $2$ for the first layer and stride $1$ for the rest. The output channel size is $1024$, $512$, $512$, and $2048$, respectively, for the four layers of Att$1$ and, moreover, each convolutional layer is followed by BN \cite{IS2015} and ReLU activation, except for the last layer which is activated by the sigmoid function instead. In contrast, both Att$2\_1$ and Att$2\_2$ consist of only one convolutional layer with kernel size $1 \times 1$, stride $1$, and output channel size same as input channel size followed by sigmoid activation.

\begin{figure}[htb]
\begin{center}
\includegraphics[width=0.9\linewidth]{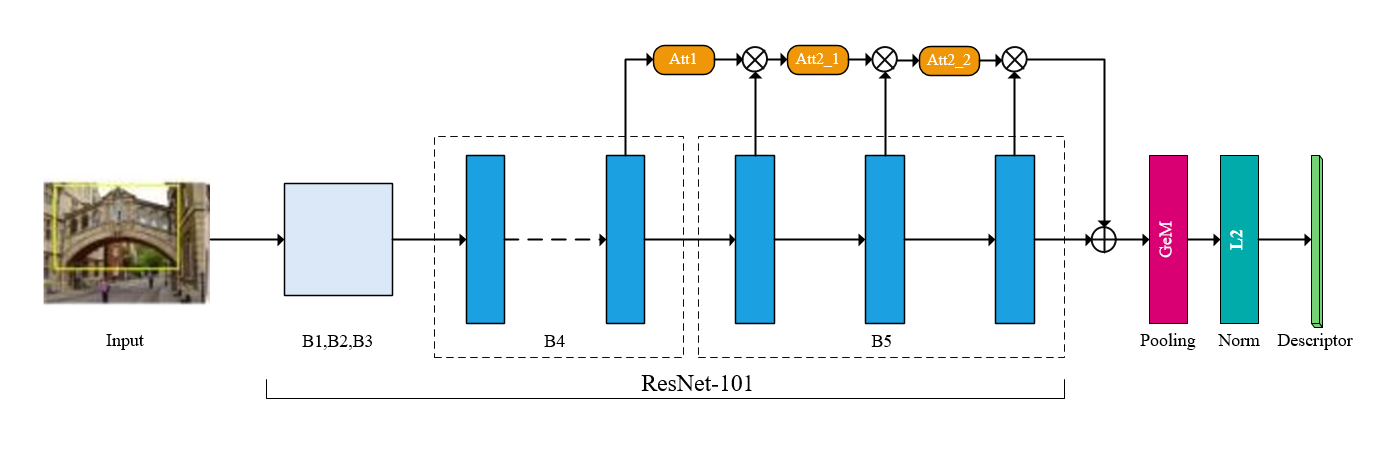}
\end{center}
\caption{A general overview of our proposed method.}
\label{fig:overall_archi}
\end{figure}

During the feedforward process, Att$1$ is applied to $\mathcal{X}_{4, 23}$ producing attention maps $\mathcal{A}_{4, 23}$, which is then combined with $\mathcal{X}_{5, 1}$ by the Hadamard product (denoted $\otimes$). Likewise, Att$2\_1$ is applied to $\mathcal{A}_{4, 23} \otimes \mathcal{X}_{5, 1}$ producing $\mathcal{A}_{5, 1}$ and Att$2\_2$ is applied to $\mathcal{A}_{5, 1} \otimes \mathcal{X}_{5, 2}$ producing $\mathcal{A}_{5, 2}$ as the output of the attention branch. The final output of the network applies attention residual learning as in \cite{WJQYLZWT2017} and produces feature maps $\mathcal{X}$ given by $\mathcal{X} = \mathcal{X}_{5, 3} + \mathcal{A}_{5, 2} \otimes \mathcal{X}_{5, 3}$, followed by GeM pooling and $\ell^2$-normalization forming a compact $2048$-D vector as the AGeM descriptor of the input image. The overall architecture is illustrated in Figure \ref{fig:overall_archi}.

\subsection{Loss Function and Whitening}

In one of the very first fine-tuning approaches proposed by Babenko {\em et al.} \cite{BSCL2014}, models pre-trained on the ImageNet dataset were re-trained on an external set of labelled landmark images using a classification loss. Later on, Gordo {\em et al.} \cite{GARL2017} showed that the results could be further improved by using a three-stream triplet loss to learn image representations that are well-suited for the retrieval task. Related to that, there is the two-stream contrastive loss used by Radenovi\'{c} {\em et al.} in \cite{RTC2016, RTC2018} given by
\begin{equation}\label{eqn:cont_loss}
L(i, j) = \begin{cases}
\frac{1}{2}\|F(i) - F(j)\|^2, & Y(i, j) = 1\\
\frac{1}{2}(\max\{0, \tau - \|F(i) - F(j)\|\})^2, & Y(i, j) = 0
\end{cases}
\end{equation}
where each training input consists of a pair of images $(i, j)$ and a label $Y(i, j) \in \{0, 1\}$ with $Y(i, j) = 1$ if $i$ and $j$ are matching and $0$ otherwise, and $\tau$ is a margin hyperparameter.

Another important component one often adds to the design of a retrieval pipeline is the PCA and whitening of image descriptors. On one hand, Gordo {\em et al.} \cite{GARL2017} proposed to append a fully connected layer (with bias) at the end of the network after the pooling layer so that PCA is learned together with the CNN. In contrast, Radenovi\'{c} {\em et al.} \cite{RTC2016, RTC2018} preferred to rather learn the discriminative whitening as a post-processing step after the fine-tuning of the CNN is finished. Since our work is closely related to that of GeM, we shall use the contrastive loss to first optimize the CNN weights, and then perform the whitening transform using the available training pairs.

\section{EXPERIMENTS}\label{sec:experim}

\subsection{Preliminary Results}

\noindent
{\bf Datasets.} For fair comparison, we will be using the same training set as \cite{RTC2016, RTC2018} which consists of around 120k images grouped into tuples, where each tuple consists of a query image, one hard positive image matching the query, and a pool of hard negative images. On the other hand, we test our method on the revised version of the Oxford5k \cite{PCISZ2007} and Paris6k \cite{PCISZ2008} retrieval benchmark datasets, referred to as ROxford5k and RParis6k \cite{RITAC2018}, and evaluate according to the Medium and Hard protocals.

\noindent
{\bf Implementation details.} We use PyTorch \cite{PGCCYDLDAL2017} for fine-tuning the network. The weights of ResNet-101 are initialized by model pre-trained on the ImageNet dataset and the pooling parameter is initialized at $p = 2.92$. We use Adam \cite{KB2015} as optimizer with initial learning rate $\ell_0 = 10^{-6}$, exponential decay $\ell_0\exp(-0.01k)$ over epoch $k$, momentum $0.9$, weight decay $10^{-4}$, and contrastive loss with margin $0.85$. For $p$ and attention layers, the initial learning rates are set to be $10^{-5}$ and $10^{-3}$. Training is done for at most 60 epochs each containing 2,000 tuples with batch size 10. Every tuple consists of 1 query image, 1 positive image, and 5 negative images selected from a pool of 20,000 negative images. All images are resized so that the longer side has size 512 while keeping the original aspect ratio. The whole process takes about 30 hours on a single NVIDIA Tesla P100 GPU with 16 GB of memory.

\noindent
{\bf Multi-scale representation.} During test time, we resize the input images to 1,024 pixels in the longer side and adopt the multi-scale scheme by extracting descriptors at different scales with scaling factors $1$, $\frac{1}{\sqrt{2}}$, and $\frac{1}{2}$. These descriptors are then combined into a single vector by average pooling followed by $\ell^2$-normalization. For GeM and AGeM descriptors, generalized mean pooling is used instead, where the pooling parameter is set to be the value learned during fine-tuning of the network.

\begin{figure}[htb]
\begin{center}
\includegraphics[width=0.9\linewidth]{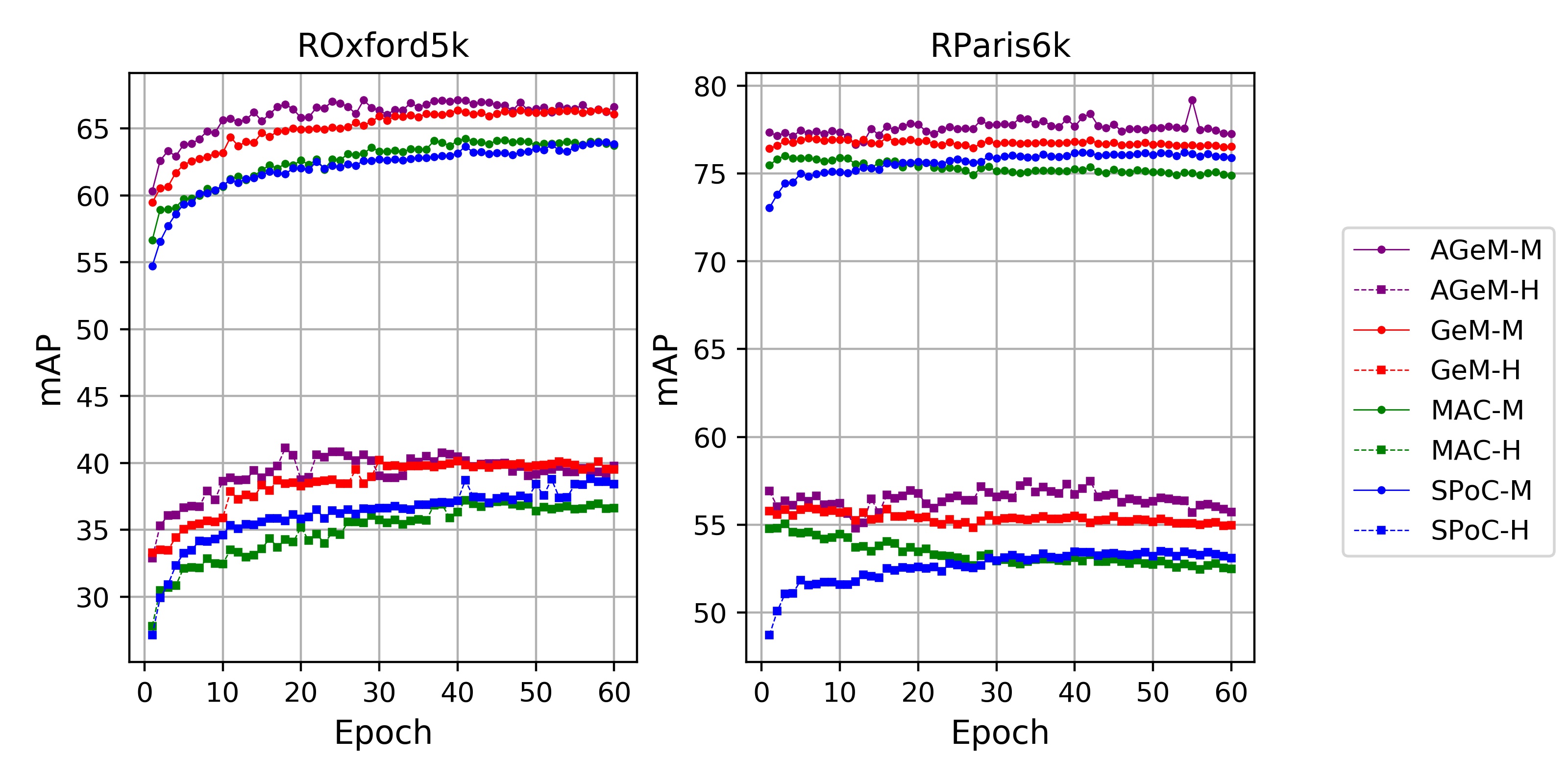}
\end{center}
\caption{Performance (mAP) comparison of AGeM, GeM, MAC, and SPoC descriptors on ROxford5k and RParis6k using Medium (-M) and Hard (-H) evaluation protocals.}
\label{fig:train_results}
\end{figure}

In Figure \ref{fig:train_results}, we present the evaluation results using our AGeM descriptors together with GeM, MAC, and SPoC descriptors on ROxford5k and RParis6k as training progresses. All descriptors are accompanied by multi-scale representation and discriminative whitening learned on the training pairs. From the plots, we observe that AGeM and GeM outperform both MAC and SPoC from beginning until the end. In general, it appears that the best performance on ROxford5k occurs around the $40$-th epoch, whereas for RParis6k the performance usually degrades in the second half of the training.

\subsection{Comparison with State-of-the-Art}

\begin{table*}[htb]
\scriptsize
\begin{center}
\begin{tabular}{|c|c|c|c|c|}
\hline
Method & ROxf-M & ROxf-H & RPar-M & RPar-H \\
\hline
\hline
$^\dagger$DIR & 60.9 & 32.4 & \textbf{78.9} & \textbf{59.4}\\
$^\dagger$GeM (MatConvNet) & 64.7 & 38.5 & 77.2 & 56.3\\
$^\dagger$GeM (PyTorch) & 65.3 & 40.0 & 76.6 & 55.2\\
$^\star$SPoC & 63.8 & 38.8 & 76.1 & 53.4\\
$^\star$MAC & 64.0 & 37.0 & 75.4 & 53.3\\
$^\star$GeM & 66.4 & 40.1 & 76.8 & 55.5\\
AGeM & \textbf{67.0} & \textbf{40.7} & 78.1 & 57.3\\
\hline
\hline
$^\dagger$DIR + $\alpha$QE & 64.8 & 36.8 & 82.7 & 65.7\\
$^\dagger$DIR + DFS & 69.0 & 44.7 & 89.5 & 80.0\\
$^\dagger$GeM + $\alpha$QE & 67.2 & 40.8 & 80.7 & 61.8\\
$^\dagger$GeM + DFS & 69.8 & 40.5 & 88.9 & 78.5\\
$^\star$SPoC + $\alpha$QE & 69.5 & 44.2 & 84.0 & 65.7\\
$^\star$SPoC + DFS & 63.8 & 38.8 & 86.4 & 73.1\\
$^\star$MAC + $\alpha$QE & 70.5 & 42.9 & 85.3 & 68.3\\
$^\star$MAC + DFS & 76.2 & 51.1 & 88.6 & 79.5\\
$^\star$GeM + $\alpha$QE & 71.6 & 45.7 & 85.7 & 69.2\\
$^\star$GeM + DFS & 78.4 & 54.0 & 88.5 & 78.8\\
AGeM + DBA + QE & 73.6 & 50.8 & 87.7 & 72.5\\
AGeM + $\beta$DBA + $\alpha$QE & 76.0 & 53.1 & 88.5 & 74.9\\
AGeM + DFS & \textbf{79.7} & 55.8 & 89.6 & 80.5\\
AGeM + $\beta$DBA + $\alpha$QE + DFS & 79.4 & \textbf{58.4} & \textbf{91.3} & \textbf{82.1}\\
\hline
\hline
$^\dagger$HesAff-rSIFT-ASMK$^*$ + SP $\rightarrow$ DIR + DFS & 80.2 & 54.8 & 92.5 & 84.0\\
$^\dagger$HesAff-rSIFT-ASMK$^*$ + SP $\rightarrow$ GeM + DFS & 79.1 & 52.7 & 91.0 & 81.0\\
$^\dagger$DELF-ASMK$^*$ + SP $\rightarrow$ DIR + DFS & 75.0 & 48.3 & 90.5 & 81.2\\
\hline
\end{tabular}
\end{center}
\caption{Performance (mAP) evaluation on ROxford5k and RParis6k using Medium (-M) and Hard (-H) evaluation protocols. For methods other than AGeM, $\dagger$ denotes results from the original papers whereas $\star$ represents our reproduction using publicly available source codes.}
\label{tab:ROxfRParResults}
\end{table*}

\noindent
{\bf Query expansion.} The simplest form of query expansion (QE) \cite{CPSIZ2007} works as follows: Given a query image, an initial search is performed and the $N$ top-ranked images are retrieved. The descriptors of these $N$ images are then combined with the descriptor of the original image by average aggregation. We refer to this as average QE (AQE).

\noindent
{\bf Database augmentation.} Like QE, one can also apply the same technique on the database side by replacing every image descriptor by a combination of itself and the descriptors of its closest neighbors. This strategy is known as database augmentation (DBA) \cite{TL2009, AZ2012, GARL2017}.

\begin{figure}[htb]
\begin{center}
\includegraphics[width=0.9\linewidth]{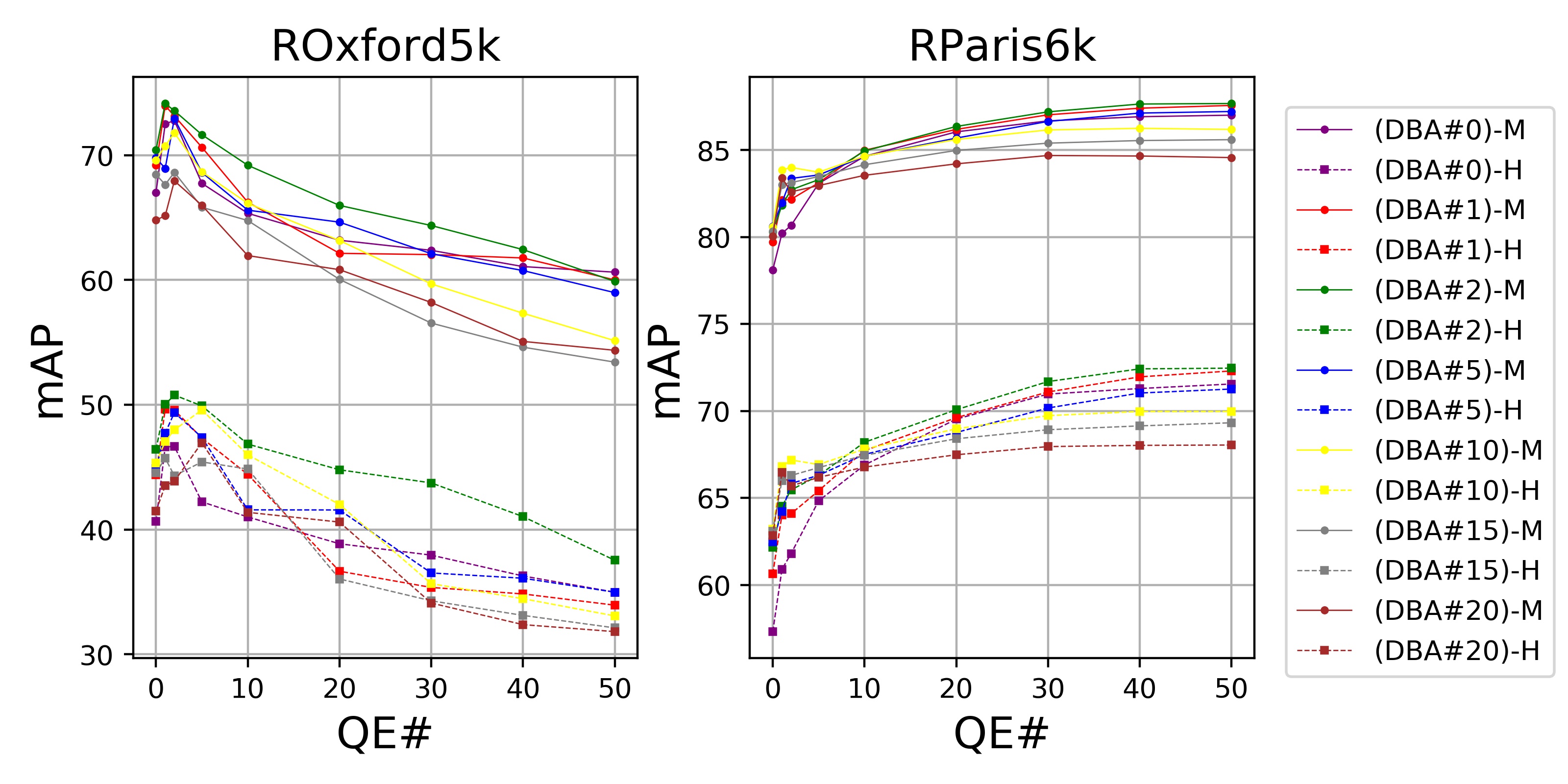}
\end{center}
\caption{mAP as a function of $\#$ of neighbors used for AQE for different values of $\#$ of neighbors used for DBA.}
\label{fig:dba_qe}
\end{figure}

In Figure \ref{fig:dba_qe}, we evaluate the effects of AQE and DBA with respect to the number of neighbors. We observe that $2$ seems to be the best number for DBA regardless of the dataset or evaluation protocal. As for QE, we suggest choosing a small number of neighbors ({\em i.e.} $1$ or $2$) for the ROxford5k and a large number of neighbors ({\em i.e.} $40$ or $50$) for the RParis6k.

\noindent
{\bf Weighted QE and DBA.} In AQE, all retrieved images have equal weight when contributing to the combined descriptor. As an alternative, $\alpha$QE was introduced in \cite{RTC2018} where for a query image $q$, the weight assigned to a top-ranked image $i$ is given by $(F(q)^{\text{T}}F(i))^\alpha$. Likewise, one can also adopt a similar weighting scheme for DBA, which we refer to as $\beta$DBA.

\begin{figure}[htb]
\begin{center}
\includegraphics[width=0.9\linewidth]{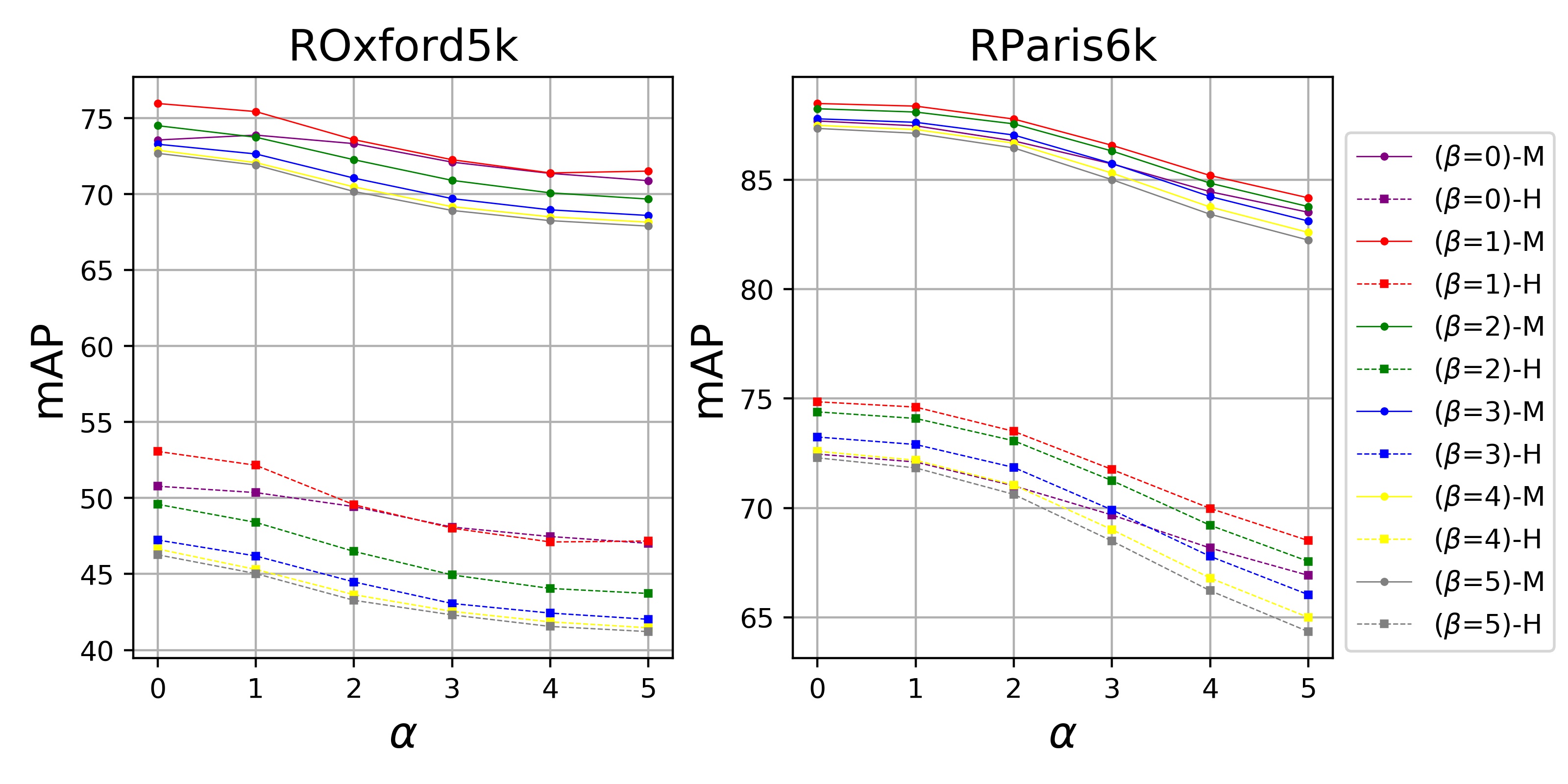}
\end{center}
\caption{mAP as a function of $\alpha$ for different values of $\beta$.}
\label{fig:dba_qe_beta_alpha}
\end{figure}

In Figure \ref{fig:dba_qe_beta_alpha}, we evaluate the effects of the exponents $\alpha$ and $\beta$ used in the weights of QE and DBA. Based on the curves in Figure \ref{fig:dba_qe_beta_alpha}, we settle for $\beta = 1$ and, perhaps surprisingly, $\alpha =0$, {\em i.e.} when $\beta$DBA is used, the best complement on the query side happens to be AQE.

In Table \ref{tab:ROxfRParResults}, we compare our proposed method against state-of-the-art. The top and middle sections correspond to methods without and with post-processing techniques, respectively, where DFS refers to the diffusion method from \cite{ITAFC2017}. One unusual observation we make is that there seems to be a non-trivial discrepancy on ROxford5k between our own implementation of GeM with DFS compared to results reported by the authors in Table $5$ of \cite{RITAC2018}. The bottom section reports the previous state-of-the-art performance taken from Table $5$ of \cite{RITAC2018}, which is not directly comparable to ours since the results are achieved by first obtaining a verified list of relevant images using local features (with methods from \cite{PCM2009, AZ2012, TAJ2015, NASWH2017, PCISZ2007}) and then starting the global-CNN-feature method from the geometrically verified images. However, even in this case, our single-method approach outperforms two of the three combinations and is on par with the last one.

\section{CONCLUSIONS}\label{sec:conclu}

In this work, we have presented an effective method to incorporate attention mechanism with CNNs for image retrieval. The proposed approach generates attention-aware features which are then combined with GeM pooling to produce a compact global descriptor with minimal overhead. The whole network can be trained end-to-end efficiently providing a substantial boost in retrieval accuracy on the challenging ROxford5k and RParis6k benchmark datasets. The result achieved by our approach outperforms or is competitive even with complex time-consuming state-of-the-art methods based on local features and spatial verification.


\bibliographystyle{IEEEbib}
\bibliography{refs}

\end{document}